\documentclass[preprint,12pt]{elsarticle}

\usepackage{amssymb}
\usepackage{amsmath,mathtools}
\usepackage{graphicx}
\usepackage[margin=1in]{geometry}
\usepackage{algorithm}
\usepackage{algorithmic}
\usepackage{multirow}
\usepackage{array}
\usepackage{booktabs} 
\usepackage{diagbox}
\usepackage{bm}
\usepackage{xcolor}

\newcommand{\best}[1]{\textbf{#1}}
\newcommand{\second}[1]{\underline{#1}}

\begin{document}

\begin{frontmatter}

\title{ITS-Mina: A Harris Hawks Optimization-Based All-MLP Framework with Iterative Refinement and External Attention for Multivariate Time Series Forecasting}

\author[1]{Pourya Zamanvaziri}
\ead{po.zamanvaziri@mail.sbu.ac.ir}
\author[1,2]{Amirhossein Sadr}
\ead{a.sadr@mail.sbu.ac.ir}
\author[2]{Aida Pakniyat}
\ead{a_pakniyat@ipm.ir}
\author[1]{Dara Rahmati}
\ead{d_rahmati@sbu.ac.ir}

\address[1]{Department of Computer Science and Engineering, Shahid Beheshti University, Iran}
\address[2]{School of Computer Science, Institute for Research in Fundamental Sciences (IPM), Iran}

\begin{abstract}
Multivariate time series forecasting plays a pivotal role in numerous real-world applications, including financial analysis, energy management, and traffic planning. While Transformer-based architectures have gained popularity for this task, recent studies reveal that simpler MLP-based models can achieve competitive or superior performance with significantly reduced computational cost. In this paper, we propose ITS-Mina, a novel all-MLP framework for multivariate time series forecasting that integrates three key innovations: (1) an iterative refinement mechanism that progressively enhances temporal representations by repeatedly applying a shared-parameter residual mixer stack, effectively deepening the model's computational capacity without multiplying the number of distinct parameters; (2) an external attention module that replaces traditional self-attention with learnable memory units, capturing cross-sample global dependencies at linear computational complexity; and (3) a Harris Hawks Optimization (HHO) algorithm for automatic dropout rate tuning, enabling adaptive regularization tailored to each dataset. Extensive experiments on six widely-used benchmark datasets demonstrate that ITS-Mina achieves state-of-the-art or highly competitive performance compared to eleven baseline models across multiple forecasting horizons. %Ablation studies and sensitivity analyses further validate the individual and collective contributions of each proposed component.
\end{abstract}


\end{frontmatter}

%%=============================================================================
%% SECTION 1: INTRODUCTION
%%=============================================================================
\section{Introduction}
\label{sec:introduction}

Time series forecasting constitutes a cornerstone of modern data-driven decision-making, with applications spanning energy consumption planning, traffic flow management, and financial analysis \cite{wu2021autoformer,zhou2022fedformer}. The multivariate setting, where multiple interrelated variables evolve jointly over time, is particularly prevalent in practice and poses unique modeling challenges. An effective forecasting model must simultaneously capture complex temporal dynamics within each variate, exploit informative cross-variate correlations, and generalize robustly across diverse forecasting horizons.

Over the past several years, Transformer-based architectures have emerged as the dominant paradigm for sequence modeling tasks, including time series forecasting \cite{vaswani2017attention,zhou2021informer,wu2021autoformer}. These models leverage self-attention mechanisms to capture long-range dependencies, and numerous variants have been proposed to address their quadratic computational complexity. Informer \cite{zhou2021informer} introduces a ProbSparse attention mechanism, Autoformer \cite{wu2021autoformer} incorporates decomposition-based auto-correlation, and FEDformer \cite{zhou2022fedformer} employs frequency-enhanced attention. Despite these advances, recent studies have revealed a counter-intuitive finding: simple linear models can match or even surpass the forecasting accuracy of sophisticated Transformer architectures on several commonly used benchmarks \cite{zeng2023dlinear}. This observation has prompted the community to reconsider the necessity of attention mechanisms for time series forecasting and to explore lightweight alternatives.

In response to this paradigm shift, MLP-based architectures have attracted considerable attention. Building on the MLP-Mixer concept from computer vision \cite{tolstikhin2021mlpmixer}, recent work has demonstrated that alternating fully-connected layers along time and feature dimensions can effectively capture both temporal patterns and cross-variate information \cite{chen2023tsmixer}. Channel-independent variants such as PatchTST \cite{nie2023patchtst} have further reinforced the competitiveness of simple architectures. However, existing MLP-based models for time series forecasting still face three fundamental limitations. First, they apply a fixed number of transformations in a single forward pass, lacking the capacity for iterative refinement that could progressively improve predictions for complex patterns. Second, they do not incorporate attention-like mechanisms to model global context across the training distribution, relying solely on local feature interactions. Third, critical hyperparameters such as dropout rates are typically set through manual grid search, which is both computationally expensive and suboptimal.

To address these limitations, we propose ITS-Mina (\textbf{I}terative \textbf{T}ime \textbf{S}eries \textbf{Mi}xer with External Atte\textbf{n}tion and H\textbf{a}rris Hawks Optimization), a novel all-MLP framework for multivariate time series forecasting. ITS-Mina builds upon the time-mixing and feature-mixing paradigm and extends it with three key innovations:

\begin{itemize}
    \item \textbf{Iterative Refinement via Shared-Parameter Mixer Loops:} We introduce an iterative refinement mechanism in which a depth-$M$ residual mixer stack is composed $N$ times with tied weights. Each round reapplies the same time--feature alternating MLP blocks, deepening the effective transformation while keeping the parameter budget of a single stack.
    
    \item \textbf{External Attention for Efficient Global Context:} We replace self-attention with an external attention module that employs two learnable memory matrices. This design captures global inter-sample correlations with $O(dSL)$ complexity (where $S$ is a fixed memory size), compared to the $O(dL^2)$ complexity of self-attention, while acting as an implicit regularizer through shared memory across samples.
    
    \item \textbf{HHO-Based Optimization:} We employ the Harris Hawks Optimization algorithm to automatically search for the optimal dropout rate, leveraging its adaptive exploration-exploitation balance to navigate the regularization landscape efficiently.
\end{itemize}

We evaluate ITS-Mina on six widely-used benchmark datasets---Traffic, Electricity, ETTh1, ETTh2, ETTm1, and ETTm2---against eleven competitive baselines spanning Transformer-based, patch-based, and MLP-based architectures. Experimental results demonstrate that ITS-Mina achieves state-of-the-art performance on the majority of dataset-horizon combinations and remains highly competitive on the remainder. 
%Comprehensive ablation studies confirm the contribution of each proposed component, and sensitivity analyses reveal the influence of key hyperparameters on forecasting accuracy.

The remainder of this paper is organized as follows: Section~\ref{sec:related_work} reviews related work. Section~\ref{sec:background} presents the background on iterative refinement, external attention, and HHO. Section~\ref{sec:method} details the proposed ITS-Mina framework. Section~\ref{sec5} reports experimental results, and Section~\ref{sec:conclusion} concludes the paper.

%%=============================================================================
%% SECTION 2: RELATED WORK
%%=============================================================================
\section{Related Work}
\label{sec:related_work}

Time series forecasting has evolved from classical statistical methods to sophisticated deep learning architectures. In the following, we review three major categories of modern approaches that are most relevant to our work.

\subsection{\textbf{Transformer-based time series models}}
\label{subsec:rw_transformer}

The success of the Transformer architecture in natural language processing \cite{vaswani2017attention} has motivated its adoption for time series forecasting, owing to its strong ability to capture long-range dependencies. However, the quadratic computational complexity of standard self-attention poses a challenge for processing long time series sequences. To address this, several efficient Transformer variants have been developed specifically for time series, such as Informer \cite{zhou2021informer}, Autoformer \cite{wu2021autoformer}, Pyraformer \cite{liu2022pyraformer}, and FEDformer \cite{zhou2022fedformer}. These models modify the attention mechanism to achieve linear or near-linear complexity, making them more suitable for long-term forecasting tasks.

\subsection{\textbf{Patch-based time series models}}
\label{subsec:rw_patch}

Instead of processing individual time points, recent patch-based approaches segment the time series into patches to improve efficiency and representation learning. PatchTST \cite{nie2023patchtst} applies a Transformer to these patches under a channel-independent assumption, where each time series is modeled separately, leading to strong performance. In contrast, CrossFormer \cite{zhang2023crossformer} employs a channel-mixing strategy within patches. While patching reduces computational cost, these models still rely on the computationally expensive self-attention mechanism at their core.

\subsection{\textbf{MLP-based time series models}}
\label{subsec:rw_mlp}

Inspired by the resurgence of Multi-Layer Perceptron (MLP) architectures in computer vision, such as MLP-Mixer \cite{tolstikhin2021mlpmixer}, recent work has explored pure MLP structures for time series. These models offer a lightweight alternative to Transformers by eliminating self-attention entirely. In time series forecasting, DLinear \cite{zeng2023dlinear} demonstrated that a simple linear model could outperform many complex Transformer-based approaches, questioning the necessity of attention mechanisms for this task. Other efforts include LightTS \cite{zhang2022lightts}, which uses MLPs with sophisticated downsampling, and MLP4Rec \cite{li2022mlp4rec}, an MLP-based model for sequential recommendation that captures correlations across time and feature dimensions.

%%=============================================================================
%% SECTION 3: BACKGROUND
%%=============================================================================
\section{Background}
\label{sec:background}

This section provides the necessary background on the three foundational concepts that underpin the design of ITS-Mina: iterative refinement and adaptive computation, external attention mechanisms, and the Harris Hawks Optimization algorithm.

\subsection{\textbf{Iterative Refinement and Adaptive Computation}}
\label{subsec:bg_iterative}

\subsubsection{Universal Transformers}
\label{subsubsec:bg_ut}

Universal Transformers \cite{dehghani2018universal} were introduced as an extension of the standard Transformer architecture, employing recurrent application of transformer layers to capture both short- and long-term dependencies. The model iteratively refines representations by looping over the transformer block. However, its evaluation was primarily conducted on smaller-scale tasks, such as WMT English-German translation, and did not involve large-scale language models like GPT-2. Furthermore, the architecture lacks a predictive residual mechanism, which is a key component for stabilizing training in larger models.

\subsubsection{Adaptive Computation Time Models}
\label{subsubsec:bg_act}

Adaptive Computation Time (ACT) \cite{graves2016adaptive} was proposed to enable recurrent neural networks to dynamically determine the number of computational steps per input. While ACT introduces adaptive computation, it was mainly applied to simpler RNN architectures and evaluated on small-scale tasks, without integration into transformer-based models or large-scale pretraining frameworks. Additionally, ACT does not incorporate a predictive residual design, distinguishing it from our approach.

\subsubsection{Depth-Adaptive Transformers}
\label{subsubsec:bg_dat}

Depth-Adaptive Transformers \cite{elbayad2019depth} adjust the network depth dynamically based on input characteristics, allowing for adaptive inference by varying the number of applied layers per sequence. However, this method focuses primarily on inference-time adaptation rather than iterative refinement during training. Evaluations were conducted on machine translation tasks with relatively modest model sizes. Notably, the architecture does not include a predictive residual design, which is central to our proposed iterative refinement framework.

\subsubsection{Loop-Residual Neural Networks}
\label{subsubsec:bg_loop}

Loop-Residual Neural Networks \cite{ng2024loop} introduced a state-of-the-art mechanism for iterative refinement in transformer architectures, particularly targeting large-scale language models. The core design iteratively loops over a subset of transformer blocks while predicting and adding residuals at each iteration, enabling progressive refinement of hidden states without increasing parameter count. This approach effectively extends computational depth and model capacity during inference while maintaining training stability through residual connections. Unlike prior adaptive computation methods, this method demonstrates improved performance through longer inference times without requiring additional parameters or training data \cite{ng2024loop}.

\subsection{\textbf{External Attention Mechanism}}
\label{subsec:bg_external}

Attention mechanisms have become integral components of modern deep learning architectures, enabling models to selectively focus on informative features and capture long-range dependencies \cite{vaswani2017attention}. Classical self-attention, popularized by the Transformer architecture, computes similarity relationships within the input itself, yielding strong representational capacity but at the cost of quadratic computational and memory complexity with respect to sequence length. This limitation has motivated substantial research into more efficient alternatives, particularly for large-scale tasks involving lengthy sequences or high-resolution data \cite{zhou2021informer}.

External attention mechanisms represent a prominent direction in this effort \cite{guo2022beyond}. Unlike self-attention, which relies solely on internal pairwise interactions, external attention introduces a set of learnable or memory-based external vectors that serve as a mediating representation. The model projects the input onto this external memory, performs attention within the compressed space, and then maps the result back to the original feature dimension. This design significantly reduces complexity by decoupling attention computation from the size of the input sequence, enabling linear-time operations while preserving the model's ability to capture global contextual information.

Another advantage of external attention is its enhanced generalization and stability. By utilizing fixed-size external memory units, the mechanism mitigates overfitting to specific sample-level correlations---an issue that can arise in self-attention when input-input interactions are overly adaptive \cite{guo2022beyond}. The external memory acts as a form of regularization, encouraging the model to learn more global structural patterns. Additionally, external attention is flexible in implementation: memory units can be static, learnable, or dynamically updated depending on the task, allowing seamless integration into convolutional, recurrent, and Transformer-based backbones.

\subsection{\textbf{Harris Hawks Optimization Algorithm}}
\label{subsec:bg_hho}

The Harris Hawks Optimization (HHO) algorithm is a nature-inspired metaheuristic introduced by Heidari et al. \cite{heidari2019hho}, motivated by the cooperative hunting strategies of Harris hawks (Parabuteo unicinctus). These birds exhibit sophisticated group-hunting behaviors, characterized by dynamic role switching, surprise pounce tactics, and coordinated pursuit of prey. HHO abstracts these behaviors into a population-based optimization framework capable of navigating complex, high-dimensional search landscapes.

HHO operates through two primary phases: exploration and exploitation, with transitions driven by a simulated ``escape energy'' of the prey. In the exploration phase, hawks diversify their search by perching in various positions and engaging in random or guided movements to probe the solution space. As the search progresses and the prey's escape energy decreases, the algorithm shifts toward exploitation, where hawks execute more focused attack maneuvers. These maneuvers include soft besiege, hard besiege, and their respective variants with progressive rapid dives, reflecting how real hawks adapt their strategy according to the prey's behavior and level of exhaustion.

One of the key strengths of HHO lies in its adaptive transition mechanism, which balances diversification and intensification without requiring user-defined switching thresholds. This makes the algorithm inherently robust in handling multimodal and nonconvex optimization problems \cite{heidari2019hho}. Moreover, by integrating stochastic components, HHO enhances its ability to avoid premature convergence and local optima---challenges commonly faced by classical optimization methods.

%%=============================================================================
%% SECTION 4: PROPOSED METHOD
%%=============================================================================
\section{Proposed Method: ITS-Mina}
\label{sec:method}

In this section, we present ITS-Mina, our proposed framework for multivariate time series forecasting. We first provide an overview of the complete architecture, then describe each of its three core components in detail: iterative refinement with a shared-parameter mixer stack, the external attention module applied after refinement, and the HHO-based dropout optimization procedure.

\section{Proposed Method: ITS-Mina}
\label{sec:method}

In this section, we present ITS-Mina, our proposed framework for multivariate time series forecasting. We first provide an overview of the complete architecture, then describe each of its three core components in detail: iterative refinement with a shared-parameter mixer stack, the external attention module applied after refinement, and the HHO-based dropout optimization procedure.

\subsection{\textbf{Overview}}
\label{subsec:overview}

\begin{figure}[t]
\centering
\includegraphics[width=\textwidth]{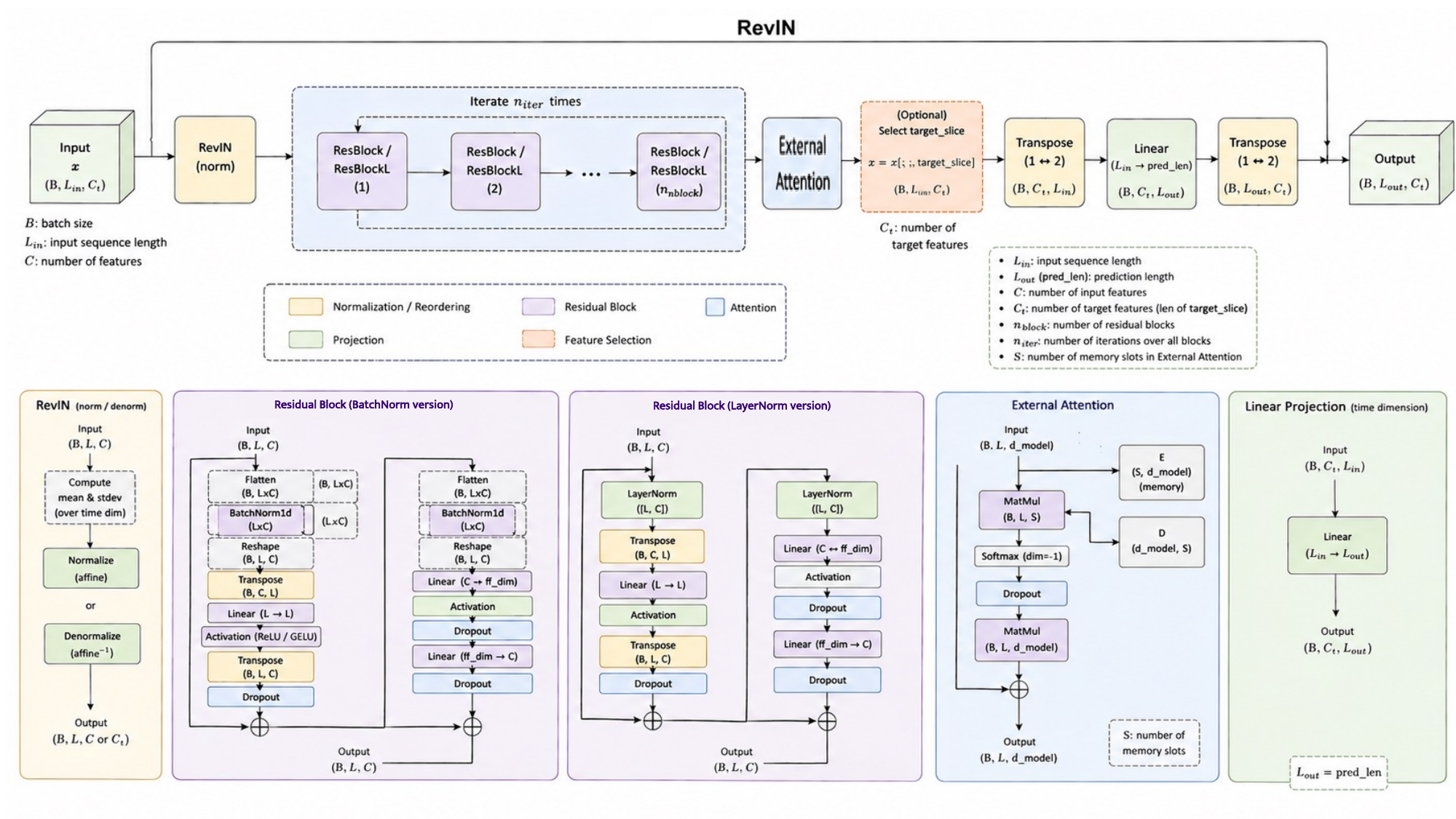}%
\caption{Overall architecture of ITS-Mina. The input is instance-normalized, refined by $N$ applications of a shared depth-$M$ residual mixer stack, augmented by external attention, and mapped to the forecast horizon with inverse normalization.}
\label{fig:overview}
\end{figure}

Let $\mathbf{X} \in \mathbb{R}^{L \times C}$ denote the input multivariate time series, where $L$ is the lookback window length and $C$ is the number of features (variates). The goal is to predict the future values $\hat{\mathbf{Y}} \in \mathbb{R}^{T \times C_{\text{out}}}$, where $T$ is the forecast horizon and $C_{\text{out}}$ is the number of predicted channels (in multivariate-to-multivariate settings $C_{\text{out}}=C$; in multivariate-to-univariate settings $C_{\text{out}}=1$ after selecting the target variate). ITS-Mina processes the input through four conceptual stages, as illustrated in Fig.~\ref{fig:overview}:

\textbf{Stage 0 -- Instance-wise normalization.} The input is first mapped to a normalized space by an instance-wise map $\Phi(\cdot)$ with learnable affine parameters, so that subsequent mixing operates on scale-stabilized representations; the inverse map $\Phi^{-1}(\cdot)$ is applied after the temporal readout in Eq.~\eqref{eq:temporal_proj}.

\textbf{Stage 1 -- Iterative refinement via a shared mixer stack.} Let $\mathbf{H}^{(0)}=\Phi(\mathbf{X})$. For $n=1,\ldots,N$, the same depth-$M$ residual mixer stack $F_{\theta}$ is applied:
\begin{equation}
\label{eq:time_mixing}
\mathbf{H}^{(n)} = F_{\theta}\!\left(\mathbf{H}^{(n-1)}\right), \quad n = 1, \ldots, N,
\end{equation}
where $F_{\theta} = f_{\theta_M} \circ \cdots \circ f_{\theta_1}$ is the composition of $M$ residual blocks $\{f_{\theta_m}\}_{m=1}^{M}$. Each $f_{\theta_m}$ is residual mixer and alternates \emph{time mixing} and \emph{feature mixing}, each preceded by normalization and followed by a residual add-back, with pointwise nonlinearity $\sigma(\cdot)$ and dropout $\mathrm{Drop}_{p_d}(\cdot)$ at training time.

Concretely, fix an outer round and write $\mathbf{G}^{(m-1)} \in \mathbb{R}^{L \times C}$ for the tensor entering block $m$ (so $\mathbf{G}^{(0)}$ is the state at the start of that round). Let $\mathrm{Norm}_1^{(m)}$ and $\mathrm{Norm}_2^{(m)}$ denote the two normalization operators inside the block (e.g., layer or batch normalization over the appropriate axes). The time-mixing sublayer maps $\mathbf{S}^{(m)} = \mathrm{Norm}_1^{(m)}\!\left(\mathbf{G}^{(m-1)}\right)$ to a residual along time via weights $\mathbf{W}_t^{(m)} \in \mathbb{R}^{L \times L}$ and bias $\mathbf{b}_t^{(m)} \in \mathbb{R}^{L}$ shared across all variates:
\begin{equation}
\label{eq:mixer_time}
\mathbf{R}_t^{(m)} = \mathrm{Drop}_{p_d}\!\left(\sigma\!\left(\mathbf{W}_t^{(m)} \mathbf{S}^{(m)} + \mathbf{b}_t^{(m)} \mathbf{1}_{C}^{\top}\right)\right),
\qquad
\mathbf{T}^{(m)} = \mathbf{G}^{(m-1)} + \mathbf{R}_t^{(m)}.
\end{equation}
Here each column $\mathbf{S}^{(m)}_{:,c}$ is an $L$-vector of variate $c$ across time; multiplying by $\mathbf{W}_t^{(m)}$ applies the \emph{same} linear map to every $c$, which is the channel-independent time-mixing design advocated in TSMixer \cite{chen2023tsmixer}. The feature-mixing sublayer then normalizes $\mathbf{T}^{(m)}$, applies a two-layer per-time-step MLP with hidden width $d_h$, and adds the result back:
\begin{equation}
\label{eq:mixer_feat}
\mathbf{Q}^{(m)} = \mathrm{Norm}_2^{(m)}\!\left(\mathbf{T}^{(m)}\right), \quad
\mathbf{H}_f^{(m)} = \sigma\!\left(\mathbf{Q}^{(m)} \mathbf{W}_{f1}^{(m)} + \mathbf{1}_{L} (\mathbf{b}_{f1}^{(m)})^{\top}\right),
\end{equation}
\begin{equation}
\label{eq:mixer_feat_out}
\mathbf{R}_f^{(m)} = \mathrm{Drop}_{p_d}\!\left(\mathbf{H}_f^{(m)} \mathbf{W}_{f2}^{(m)} + \mathbf{1}_{L} (\mathbf{b}_{f2}^{(m)})^{\top}\right),
\qquad
\mathbf{G}^{(m)} = \mathbf{T}^{(m)} + \mathbf{R}_f^{(m)},
\end{equation}
with $\mathbf{W}_{f1}^{(m)} \in \mathbb{R}^{C \times d_h}$, $\mathbf{W}_{f2}^{(m)} \in \mathbb{R}^{d_h \times C}$, and biases $\mathbf{b}_{f1}^{(m)} \in \mathbb{R}^{d_h}$, $\mathbf{b}_{f2}^{(m)} \in \mathbb{R}^{C}$. Thus $f_{\theta_m}(\mathbf{G}^{(m-1)}) := \mathbf{G}^{(m)}$, and the learnable parameters of block $m$ are $\theta_m = \{\mathbf{W}_t^{(m)}, \mathbf{b}_t^{(m)}, \mathbf{W}_{f1}^{(m)}, \mathbf{b}_{f1}^{(m)}, \mathbf{W}_{f2}^{(m)}, \mathbf{b}_{f2}^{(m)}\}$ together with the affine parameters inside $\mathrm{Norm}_1^{(m)}$ and $\mathrm{Norm}_2^{(m)}$. One full application of the stack is $F_{\theta}(\mathbf{G}^{(0)}) = \mathbf{G}^{(M)}$.

The same $\theta = \{\theta_1,\ldots,\theta_M\}$ is reused at every outer round $n$ in Eq.~\eqref{eq:time_mixing}, so the network evaluates $N M$ mixer blocks in total while storing only one copy of each $\theta_m$; increasing $N$ deepens computation under \emph{weight tying} rather than by widening the parameter budget.

\textbf{Stage 2 -- External attention.} After the $N$-th refinement round, the representation $\mathbf{H}^{(N)}$ is passed through the external attention module described in Section~\ref{subsec:external}, which adds a slot-conditioned residual correction and thereby injects global, dataset-level context before forecasting heads are applied.

\textbf{Stage 3 -- Temporal projection and denormalization.} Let $\mathbf{Z}$ denote the post-attention tensor (sliced to the target channels when $C_{\text{out}}<C$). The forecast is obtained by a linear readout along time and mapped back to the original scale by the inverse $\Phi^{-1}(\cdot)$ of the instance normalization:
\begin{equation}
\label{eq:temporal_proj}
\tilde{\mathbf{Y}} = \left(\mathbf{W}_p \, \mathbf{Z}^{\top} + \mathbf{b}_p\right)^{\!\top},
\qquad
\hat{\mathbf{Y}} = \Phi^{-1}(\tilde{\mathbf{Y}}),
\end{equation}
where $\mathbf{W}_p \in \mathbb{R}^{T \times L}$ and $\mathbf{b}_p \in \mathbb{R}^{T}$ are learnable parameters. The model is trained end-to-end by minimizing the mean squared error (MSE) loss:
\begin{equation}
\label{eq:loss}
\mathcal{L} = \frac{1}{T \cdot C_{\text{out}}} \sum_{t=1}^{T} \sum_{c=1}^{C_{\text{out}}} \left(Y_{t,c} - \hat{Y}_{t,c}\right)^2.
\end{equation}

The overall forward pass of ITS-Mina is summarized in Algorithm~\ref{alg:overall}.

\begin{algorithm}[t]
\caption{ITS-Mina Forward Pass}
\label{alg:overall}
\begin{algorithmic}
\REQUIRE Input time series $\mathbf{X} \in \mathbb{R}^{L \times C}$; outer refinement rounds $N$; mixer depth $M$ per round; dropout rate $p_d$; instance map $\Phi$ and its inverse $\Phi^{-1}$; temporal projection $(\mathbf{W}_p, \mathbf{b}_p)$
\ENSURE Forecast $\hat{\mathbf{Y}} \in \mathbb{R}^{T \times C_{\text{out}}}$
\STATE $\mathbf{H} \leftarrow \Phi(\mathbf{X})$
\FOR{$n = 1$ to $N$}
    \FOR{$m = 1$ to $M$}
        \STATE $\mathbf{H} \leftarrow \text{ResidualMixerBlock}\!\left(\mathbf{H};\,\theta_m,\,p_d\right)$ \hfill $\triangleright$ time-then-feature mixing with internal residuals (Algorithm~\ref{alg:iterative})
    \ENDFOR
\ENDFOR
\STATE $\mathbf{Z} \leftarrow \text{ExternalAttention}(\mathbf{H})$ \hfill $\triangleright$ Algorithm~\ref{alg:external}
\STATE restrict $\mathbf{Z}$ to target channels if $C_{\text{out}} < C$
\STATE $\tilde{\mathbf{Y}} \leftarrow (\mathbf{W}_p \, \mathbf{Z}^{\top} + \mathbf{b}_p)^{\top}$
\STATE $\hat{\mathbf{Y}} \leftarrow \Phi^{-1}(\tilde{\mathbf{Y}})$
\RETURN $\hat{\mathbf{Y}}$
\end{algorithmic}
\end{algorithm}

%%-----------------------------------------------------------------------------
\subsection{\textbf{Addressing Iterative Refinement}}
\label{subsec:iterative}

A key limitation of conventional MLP-based time series models is their reliance on a single forward pass through independently parameterized layers. While stacking more layers increases capacity, it also proportionally increases the parameter count and the risk of overfitting, particularly on small-to-medium-sized datasets that are common in time series benchmarks.

ITS-Mina addresses this limitation by \emph{weight tying across refinement rounds}. The building block is a depth-$M$ residual mixer stack $F_{\theta}=f_{\theta_M}\circ\cdots\circ f_{\theta_1}$ as in Eq.~\eqref{eq:time_mixing}. Starting from the instance-normalized input $\mathbf{H}^{(0)}=\Phi(\mathbf{X})$, the same map $F_{\theta}$ is applied $N$ times,
\begin{equation}
\label{eq:iterative_update}
\mathbf{H}^{(n)} = F_{\theta}\!\left(\mathbf{H}^{(n-1)}\right), \quad n = 1, \ldots, N,
\end{equation}
which is identical to Eq.~\eqref{eq:time_mixing}. Equivalently, the representation after $N$ rounds is the $N$-fold iterate of $F_{\theta}$:
\begin{equation}
\label{eq:unrolled}
\mathbf{H}^{(N)} = \underbrace{F_{\theta} \circ F_{\theta} \circ \cdots \circ F_{\theta}}_{N\ \text{times}}\!\left(\mathbf{H}^{(0)}\right).
\end{equation}
Thus iterative refinement here deepens the \emph{effective} computation by composing the same nonlinear transformation, rather than by introducing $N$ independent copies of every layer. External attention is deferred until after these rounds (Section~\ref{subsec:overview}), so each application of $F_{\theta}$ refines temporal and cross-variate structure before global memory is injected at the end of Stage~1.

This construction has several desirable properties:

\begin{enumerate}
    \item \textbf{Parameter efficiency:} All $N$ outer rounds share the same parameter set $\theta=\{\theta_1,\ldots,\theta_M\}$. Increasing $N$ increases the number of residual-block evaluations---and hence receptive-field depth under composition---without multiplying the number of distinct weight tensors.
    
    \item \textbf{Gradient stability:} Each inner block $f_{\theta_m}$ is itself residual; composing $F_{\theta}$ and repeating it preserves short gradient paths through those internal skips, which helps training when $N$ or $M$ is moderately large \cite{he2016deep}.
    
    \item \textbf{Progressive refinement:} Repeated application of the same $F_{\theta}$ induces a dynamical system on hidden states: early rounds can emphasize coarse corrections compatible with the map's Jacobian at typical activations, while later rounds further contract representation error before the single external-attention stage. No per-round gating vectors are required; modulation arises from the nonlinearity and normalization inside $F_{\theta}$ together with depth under iteration.
\end{enumerate}

Algorithm~\ref{alg:iterative} expands one evaluation of $F_{\theta}$ into its $M$ residual mixer blocks; Algorithm~\ref{alg:overall} nests this procedure within the $N$ refinement rounds, then applies external attention and the temporal readout.

\begin{algorithm}[t]
\caption{One refinement round: depth-$M$ residual mixer stack $F_{\theta}$}
\label{alg:iterative}
\begin{algorithmic}
\REQUIRE Current state $\mathbf{H} \in \mathbb{R}^{L \times C}$; block parameters $\{\theta_m\}_{m=1}^{M}$; dropout rate $p_d$
\ENSURE Updated state $F_{\theta}(\mathbf{H}) \in \mathbb{R}^{L \times C}$
\FOR{$m = 1$ to $M$}
    \STATE $\mathbf{H} \leftarrow \text{ResidualMixerBlock}\!\left(\mathbf{H};\,\theta_m,\,p_d\right)$ \hfill $\triangleright$ time-mixing then feature-mixing with residuals and dropout
\ENDFOR
\RETURN $\mathbf{H}$
\end{algorithmic}
\end{algorithm}

%%-----------------------------------------------------------------------------
%%-----------------------------------------------------------------------------
\subsection{\textbf{Addressing External Attention}}
\label{subsec:external}

Standard self-attention computes pairwise similarity scores between all positions in the input sequence, resulting in $O(L^2)$ complexity that becomes prohibitive for long time series. Moreover, self-attention operates exclusively within a single input sample, limiting its ability to capture inter-sample correlations that may encode useful global structural patterns.

ITS-Mina employs an external attention mechanism that addresses both limitations. We adopt the same tensor layout as in Section~\ref{subsec:overview}: time steps index rows and variates index columns. Let $\mathbf{H} \in \mathbb{R}^{L \times C}$ denote the representation fed into attention---in the full model, $\mathbf{H}=\mathbf{H}^{(N)}$ after iterative refinement in Eq.~\eqref{eq:time_mixing}. External attention introduces two learnable \emph{slot} matrices $\mathbf{E} \in \mathbb{R}^{S \times C}$ and $\mathbf{V} \in \mathbb{R}^{S \times C}$, where $S$ is a fixed memory capacity (number of slots). Rows $\mathbf{E}_{s,:}$ and $\mathbf{V}_{s,:}$ act as shared external prototypes in variate space; they are \emph{input-independent} parameters trained jointly with the rest of the model, following the external-attention principle \cite{guo2022beyond}.

The computation proceeds in three steps. First, for each time index $\ell \in \{1,\ldots,L\}$, affinities between the local variate vector $\mathbf{H}_{\ell,:}$ and every slot are formed by a single linear map:
\begin{equation}
\label{eq:ext_affinity}
\hat{\mathbf{A}} = \mathbf{H} \, \mathbf{E}^{\top} \in \mathbb{R}^{L \times S},
\end{equation}
where $\hat{A}_{\ell,s}$ scores how strongly time step $\ell$ aligns with slot $s$.

Second, affinities are normalized into attention weights over the slot index (row-wise softmax):
\begin{equation}
\label{eq:ext_softmax}
A_{\ell,s} = \frac{\exp\!\left(\hat{A}_{\ell,s}\right)}{\sum_{s'=1}^{S} \exp\!\left(\hat{A}_{\ell,s'}\right)}.
\end{equation}

Third, a slot-aggregated correction is mapped back to variate dimension and added residually to $\mathbf{H}$. With $\mathbf{A} \in \mathbb{R}^{L \times S}$ and $\mathbf{V} \in \mathbb{R}^{S \times C}$,
\begin{equation}
\label{eq:ext_output}
\mathbf{R} = \mathbf{A} \, \mathbf{V} \in \mathbb{R}^{L \times C},
\qquad
\mathbf{Z} = \mathbf{H} + \mathbf{R},
\end{equation}
which defines the post-attention tensor $\mathbf{Z}$ used in Eq.~\eqref{eq:temporal_proj}.

\begin{figure}[t]
\centering
\includegraphics[width=4.5in]{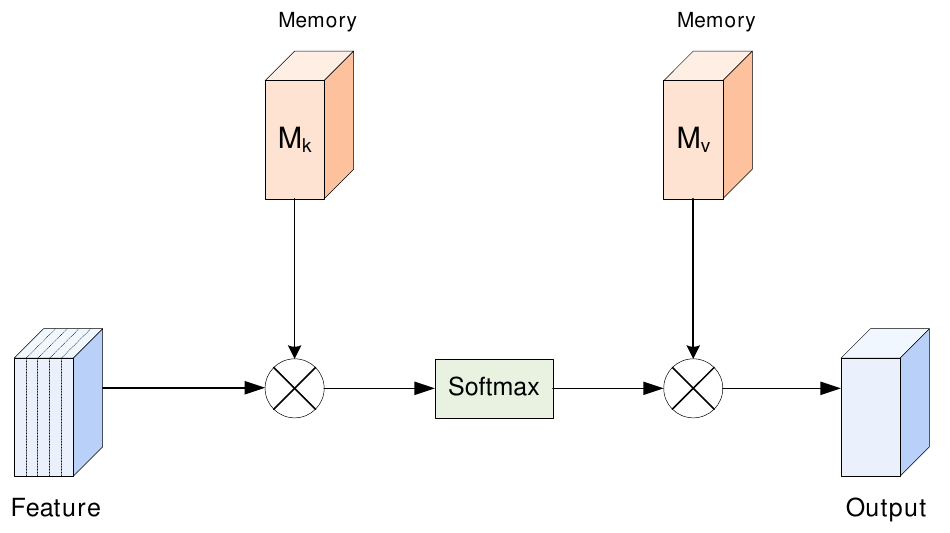}%
\caption{External attention in ITS-Mina (notation aligned with Section~\ref{subsec:overview}). Rows of $\mathbf{H} \in \mathbb{R}^{L \times C}$ attend over $S$ shared slots via $\mathbf{E}$; the mixture is mapped by $\mathbf{V}$ and added residually to yield $\mathbf{Z}$.}
\label{fig:external}
\end{figure}

The complete external attention module is illustrated in Fig.~\ref{fig:external}. Its dominant cost is two matrix products of sizes $(L \times C)(C \times S)$ and $(L \times S)(S \times C)$, i.e., $O(L C S)$, which is linear in the lookback length $L$ for fixed $C$ and $S$. This contrasts with self-attention's $O(L^2)$ dependence on the sequence length when the embedding dimension is held fixed.

A critical advantage of external attention is its implicit regularization effect. Since $\mathbf{E}$ and $\mathbf{V}$ are shared across all training windows, they encourage representations that transfer across samples rather than memorizing window-specific idiosyncrasies. Writing $\{\omega_1, \ldots, \omega_P\}$ for a sequence of training batches (or windows), a schematic gradient update is
\begin{equation}
\label{eq:memory_update}
\mathbf{E}^{(i)} = \mathbf{E}^{(i-1)} - \eta \frac{\partial \mathcal{L}(\omega_i)}{\partial \mathbf{E}^{(i-1)}}, \quad
\mathbf{V}^{(i)} = \mathbf{V}^{(i-1)} - \eta \frac{\partial \mathcal{L}(\omega_i)}{\partial \mathbf{V}^{(i-1)}},
\end{equation}
where $\eta$ is the learning rate and $\mathcal{L}(\omega_i)$ is the loss on batch $\omega_i$. After optimization, $\mathbf{E}$ and $\mathbf{V}$ encode recurring patterns seen across the training distribution, complementing the per-window mixing in $\mathbf{H}^{(N)}$.

\begin{algorithm}[t]
\caption{External Attention Module}
\label{alg:external}
\begin{algorithmic}
\REQUIRE Representation $\mathbf{H} \in \mathbb{R}^{L \times C}$; learnable slots $\mathbf{E}, \mathbf{V} \in \mathbb{R}^{S \times C}$
\ENSURE Post-attention representation $\mathbf{Z} \in \mathbb{R}^{L \times C}$
\STATE $\hat{\mathbf{A}} \leftarrow \mathbf{H} \, \mathbf{E}^{\top}$ \hfill $\triangleright$ Affinities: $\mathbb{R}^{L \times S}$
\STATE $\mathbf{A} \leftarrow \text{Softmax}(\hat{\mathbf{A}})$ applied independently to each row (weights over the slot index)
\STATE $\mathbf{R} \leftarrow \mathbf{A} \, \mathbf{V}$ \hfill $\triangleright$ Correction: $\mathbb{R}^{L \times C}$
\STATE $\mathbf{Z} \leftarrow \mathbf{H} + \mathbf{R}$
\RETURN $\mathbf{Z}$
\end{algorithmic}
\end{algorithm}

The overall architecture of external attention module is summarized in Algorithm~\ref{alg:external}.

%%-----------------------------------------------------------------------------
\subsection{\textbf{Addressing Harris Hawks Optimization}}
\label{subsec:hho}

Dropout \cite{srivastava2014dropout} is a widely-used regularization technique that randomly zeroes out a fraction of neurons during training, preventing co-adaptation and improving generalization. The dropout rate $p_d \in [0, 1)$ is a critical hyperparameter: too low a value provides insufficient regularization, while too high a value impedes the model's learning capacity. Traditional approaches rely on grid search or random search over a discrete set of candidates, which is both computationally expensive and unlikely to find the global optimum in the continuous search space.

We formulate dropout rate optimization as a continuous optimization problem and employ the Harris Hawks Optimization (HHO) algorithm \cite{heidari2019hho} to solve it. A population of $H$ hawks (candidate solutions) searches the dropout rate space $[0, 0.5]$, where each hawk $i$ maintains a candidate dropout rate $p_d^{(i)}(t)$ at iteration $t$. The fitness of each candidate is evaluated by training the ITS-Mina model with the corresponding dropout rate for a fixed number of epochs and measuring the validation MSE:
\begin{equation}
\label{eq:fitness}
\mathcal{F}(p_d^{(i)}) = \text{MSE}_{\text{val}}\!\left(\mathcal{M}_\Theta(p_d^{(i)})\right),
\end{equation}
where $\mathcal{M}_\Theta(p_d^{(i)})$ denotes the ITS-Mina model with structural parameters $\Theta$ (optimized by Optuna) and dropout rate $p_d^{(i)}$. The rabbit (prey) position $p_{\text{rabbit}}(t)$ corresponds to the best dropout rate found so far, i.e., the one yielding the lowest validation MSE.

\textbf{Escape energy.} The transition between exploration and exploitation is governed by the prey's escape energy:
\begin{equation}
\label{eq:escape_energy}
E(t) = 2 E_0 \left(1 - \frac{t}{T_{\max}}\right),
\end{equation}
where $E_0 \in [-1, 1]$ is a random initial energy and $T_{\max}$ is the maximum number of HHO iterations. When $|E| \geq 1$, the algorithm is in exploration mode; when $|E| < 1$, it switches to exploitation.

\textbf{Exploration phase} ($|E| \geq 1$). Hawks diversify their search to broadly explore the dropout rate space:
\begin{equation}
\label{eq:exploration}
p_d^{(i)}(t+1) =
\begin{cases}
p_{\text{rand}}(t) - r_1 \left| p_{\text{rand}}(t) - 2r_2 \, p_d^{(i)}(t) \right|, & \text{if } q \geq 0.5, \\[6pt]
\left(p_{\text{rabbit}}(t) - \bar{p}(t)\right) - r_3 \left(\text{LB} + r_4 \left(\text{UB} - \text{LB}\right)\right), & \text{if } q < 0.5,
\end{cases}
\end{equation}
where $p_{\text{rand}}(t)$ is a randomly selected hawk's position, $\bar{p}(t) = \frac{1}{H}\sum_{i=1}^{H} p_d^{(i)}(t)$ is the population mean, $r_1, r_2, r_3, r_4, q \sim \mathcal{U}(0,1)$ are random numbers, and $\text{LB} = 0$, $\text{UB} = 0.5$ define the search bounds.

\textbf{Exploitation phase} ($|E| < 1$). Hawks execute focused attack maneuvers with four strategies depending on the escape energy $E$ and a random parameter $r \sim \mathcal{U}(0,1)$:

\noindent\emph{Soft besiege} ($r \geq 0.5$, $|E| \geq 0.5$):
\begin{equation}
\label{eq:soft_besiege}
p_d^{(i)}(t+1) = p_{\text{rabbit}}(t) - E(t) \left| J \cdot p_{\text{rabbit}}(t) - p_d^{(i)}(t) \right|,
\end{equation}
where $J = 2(1 - r_5)$ with $r_5 \sim \mathcal{U}(0,1)$ represents the prey's random jump strength.

\noindent\emph{Hard besiege} ($r \geq 0.5$, $|E| < 0.5$):
\begin{equation}
\label{eq:hard_besiege}
p_d^{(i)}(t+1) = p_{\text{rabbit}}(t) - E(t) \left| p_{\text{rabbit}}(t) - p_d^{(i)}(t) \right|.
\end{equation}

\noindent\emph{Soft besiege with progressive rapid dives} ($r < 0.5$, $|E| \geq 0.5$):
\begin{equation}
\label{eq:soft_dive}
p_d^{(i)}(t+1) =
\begin{cases}
\Delta p, & \text{if } \mathcal{F}(\Delta p) < \mathcal{F}(p_d^{(i)}(t)), \\
\Delta p + r_6 \cdot \text{L\'{e}vy}(1), & \text{otherwise},
\end{cases}
\end{equation}
where $\Delta p = p_{\text{rabbit}}(t) - E(t) \left| J \cdot p_{\text{rabbit}}(t) - p_d^{(i)}(t) \right|$ and $r_6 \sim \mathcal{U}(0,1)$.

\noindent\emph{Hard besiege with progressive rapid dives} ($r < 0.5$, $|E| < 0.5$):
\begin{equation}
\label{eq:hard_dive}
p_d^{(i)}(t+1) =
\begin{cases}
\Delta p', & \text{if } \mathcal{F}(\Delta p') < \mathcal{F}(p_d^{(i)}(t)), \\
\Delta p' + r_6 \cdot \text{L\'{e}vy}(1), & \text{otherwise},
\end{cases}
\end{equation}
where $\Delta p' = p_{\text{rabbit}}(t) - E(t) \left| p_{\text{rabbit}}(t) - \bar{p}(t) \right|$.

In all cases, after updating, the dropout rate is clamped to $[\text{LB}, \text{UB}]$. The complete HHO-based dropout optimization procedure is presented in Algorithm~\ref{alg:hho}.

\begin{algorithm}[t]
\caption{HHO-Based Dropout Rate Optimization}
\label{alg:hho}
\begin{algorithmic}
\REQUIRE Population size $H$; max iterations $T_{\max}$; search bounds $[\text{LB}, \text{UB}] = [0, 0.5]$; pre-optimized model parameters $\Theta$ (via Optuna)
\ENSURE Optimal dropout rate $p_d^{*}$
\STATE Initialize hawk population $\{p_d^{(i)}(0)\}_{i=1}^{H}$ uniformly in $[\text{LB}, \text{UB}]$
\STATE Evaluate fitness $\mathcal{F}(p_d^{(i)}(0))$ for all $i$ via Eq.~(\ref{eq:fitness})
\STATE $p_{\text{rabbit}} \leftarrow \arg\min_{p_d^{(i)}} \mathcal{F}(p_d^{(i)}(0))$
\FOR{$t = 1$ to $T_{\max}$}
    \FOR{$i = 1$ to $H$}
        \STATE Compute $E(t)$ via Eq.~(\ref{eq:escape_energy})
        \IF{$|E(t)| \geq 1$}
            \STATE Update $p_d^{(i)}(t)$ via Eq.~(\ref{eq:exploration}) \hfill $\triangleright$ Exploration
        \ELSE
            \STATE Draw $r \sim \mathcal{U}(0,1)$
            \IF{$r \geq 0.5$ and $|E(t)| \geq 0.5$}
                \STATE Update via Eq.~(\ref{eq:soft_besiege}) \hfill $\triangleright$ Soft besiege
            \ELSIF{$r \geq 0.5$ and $|E(t)| < 0.5$}
                \STATE Update via Eq.~(\ref{eq:hard_besiege}) \hfill $\triangleright$ Hard besiege
            \ELSIF{$r < 0.5$ and $|E(t)| \geq 0.5$}
                \STATE Update via Eq.~(\ref{eq:soft_dive}) \hfill $\triangleright$ Soft besiege + rapid dives
            \ELSE
                \STATE Update via Eq.~(\ref{eq:hard_dive}) \hfill $\triangleright$ Hard besiege + rapid dives
            \ENDIF
        \ENDIF
        \STATE $p_d^{(i)}(t) \leftarrow \text{clamp}(p_d^{(i)}(t),\ \text{LB},\ \text{UB})$
        \STATE Evaluate $\mathcal{F}(p_d^{(i)}(t))$
    \ENDFOR
    \STATE $p_{\text{rabbit}} \leftarrow \arg\min_{p_d^{(i)}} \mathcal{F}(p_d^{(i)}(t))$
\ENDFOR
\RETURN $p_d^{*} \leftarrow p_{\text{rabbit}}$
\end{algorithmic}
\end{algorithm}

All other hyperparameters of ITS-Mina---including the number of time-mixing iterations $N$, the number of feature-mixing layers $M$, the external memory size $S$, the learning rate, the batch size, and the hidden dimension $d_h$---are optimized using the Optuna framework \cite{akiba2019optuna}, which employs a tree-structured Parzen estimator (TPE) for efficient Bayesian hyperparameter search. The overall training pipeline thus consists of two phases: (1) Optuna-based structural hyperparameter optimization, and (2) HHO-based dropout rate refinement with the structural parameters fixed.

%%=============================================================================
%% SECTION 5: EXPERIMENTAL RESULTS
%%=============================================================================
\section{Experimental Results}
\label{sec5}

This section presents a comprehensive evaluation of ITS-Mina on six widely-used multivariate time series forecasting benchmarks. We describe the experimental setup, and compare our method against state-of-the-art baselines. 
%and provide detailed analyses including sensitivity studies and ablation experiments.

%% Use \subsection commands to start a subsection.
\subsection{\textbf{Dataset Description}}
\label{subsec:exp1}

We evaluate ITS-Mina on six popular benchmark datasets that span diverse application domains. %\textbf{Weather} contains 21 meteorological features recorded at 10-minute intervals.
\textbf{Traffic} comprises 862 sensors measuring road occupancy rates at hourly intervals. \textbf{Electricity} includes hourly electricity consumption of 321 clients. The \textbf{ETT} (Electricity Transformer Temperature) datasets---ETTh1, ETTh2, ETTm1, and ETTm2---each contain 7 features recorded at hourly (h) or 15-minute (m) intervals. Table~\ref{tab:dataset_statistics} summarizes the statistics of all datasets. For all datasets, we follow the standard chronological split for training, validation, and testing \cite{wu2021autoformer,nie2023patchtst}.

\begin{table}[t]
  \centering
  \caption{Statistics of popular datasets for benchmark.}
  \label{tab:dataset_statistics}
  \footnotesize
  \begin{tabular}{lcccccc}
    \toprule
    Datasets & Traffic & Electricity & ETTh1 & ETTh2 & ETTm1 & ETTm2 \\
    \midrule
    Features & 862 & 321 & 7 & 7 & 7 & 7 \\
    Timesteps & 17544 & 26304 & 17420 & 17420 & 69680 & 69680 \\
    \bottomrule
  \end{tabular}
\end{table}

\subsection{\textbf{Evaluation Metrics}}
\label{subsec:exp2}

We adopt the two standard evaluation metrics for multivariate time series forecasting: Mean Squared Error (MSE) and Mean Absolute Error (MAE). Given ground truth $\boldsymbol{Y} \in \mathbb{R}^{T \times C}$ and prediction $\hat{\boldsymbol{Y}} \in \mathbb{R}^{T \times C}$, these are defined as:
\begin{equation}
    \text{MSE} = \frac{1}{T \cdot C} \sum_{t=1}^{T} \sum_{c=1}^{C} \left(Y_{t,c} - \hat{Y}_{t,c}\right)^2,
\label{eq:mse}
\end{equation}
\begin{equation}
    \text{MAE} = \frac{1}{T \cdot C} \sum_{t=1}^{T} \sum_{c=1}^{C} \left|Y_{t,c} - \hat{Y}_{t,c}\right|.
\label{eq:mae}
\end{equation}
Lower values indicate better forecasting performance for both metrics.

\subsection{\textbf{Parameter Settings}}
\label{subsec:params}

The structural hyperparameters of ITS-Mina are optimized using Optuna \cite{akiba2019optuna} with a TPE sampler. The search spaces are: number of time-mixing iterations $N \in \{2, 4, 6, 8\}$, number of residual blocks $M \in \{2, 4, 8\}$, external attention memory size \(S \in \{8, 16, 32, 64, 128, 256\}\), hidden dimension $d_h \in [16, 2048]$ sampled as a log-uniform integer, learning rate $\eta \in [10^{-4}, 10^{-2}]$ (log-uniform), input sequence length \(seq\_len \in \{4, 8, 16, 32, 64, 128, 256, 512, 1024\}\), batch size $B \in \{16, 32, 64\}$ and normalization type (BatchNorm or LayerNorm).

The dropout rate is subsequently optimized via HHO with population size $H = 10$ and $T_{\max} = 20$ iterations, searching within $[0, 0.5]$. Each fitness evaluation trains the model for 10 epochs and measures validation MSE.

All models are trained using the Adam optimizer \cite{kingma2015adam} with the optimized learning rate for a maximum of 100 epochs with early stopping (patience of 5 epochs based on validation loss). Experiments are conducted on NVIDIA RTX 3090 GPUs. 
%We report the mean results over 3 independent runs to ensure statistical reliability.

\subsection{\textbf{Compared Models}}
\label{subsec:exp4}

We compare ITS-Mina against eleven representative baselines spanning three architectural families:

\textbf{Transformer-based models:} Informer \cite{zhou2021informer} with ProbSparse attention; Autoformer \cite{wu2021autoformer} with decomposition-based auto-correlation; FEDformer \cite{zhou2022fedformer} with frequency-enhanced attention; Crossformer \cite{zhang2023crossformer} with cross-dimension dependency; iTransformer \cite{liu2024itransformer} with inverted Transformer design.

\textbf{MLP-based models:} DLinear \cite{zeng2023dlinear}, a simple decomposition-linear model; CI-TSMixer \cite{chen2023tsmixer}, a channel-independent variant of TSMixer with time-mixing and feature-mixing MLPs; PatchMixer \cite{li2023patchmixer}, combining patch-based representations with MLP mixing.

\textbf{Other models:} PatchTST \cite{nie2023patchtst}, a patch-based Transformer with channel independence; TimesNet \cite{wu2023timesnet}, a convolution-based model with temporal 2D variation modeling; MICN \cite{wang2023micn}, a multi-scale isometric convolution network.

For all baselines, we use the results reported in their original papers or reproduce them using official implementations with recommended hyperparameters to ensure fair comparison.

\subsection{\textbf{Comparison Results}}
\label{subsec:results}

Tables~\ref{tab:mse_comparison} and~\ref{tab:mae_comparison} present the MSE and MAE results, respectively, across all six datasets and four forecast horizons. The best result in each row is shown in \textbf{bold} and the second best is \underline{underlined}.

\begin{table}[t!]
  \centering
  \caption{MSE comparison across different horizons and datasets. The best result in each row is in \textbf{bold} and the second best is \underline{underlined}.}
  \label{tab:mse_comparison}
  \footnotesize
  \resizebox{\linewidth}{!}{
  \begin{tabular}{llcccccccccccc}
    \toprule
    \multicolumn{2}{c}{\diagbox[width=3cm,height=2.1cm]{Datasets}{Models}} &
      \rotatebox[origin=c]{90}{\textbf{ITS-Mina}} &
      \rotatebox[origin=c]{90}{CI-TSMixer} &
      \rotatebox[origin=c]{90}{PatchMixer} &
      \rotatebox[origin=c]{90}{PatchTST} &
      \rotatebox[origin=c]{90}{Crossformer} &
      \rotatebox[origin=c]{90}{iTransformer} &
      \rotatebox[origin=c]{90}{TimesNet} &
      \rotatebox[origin=c]{90}{MICN} &
      \rotatebox[origin=c]{90}{DLinear} &
      \rotatebox[origin=c]{90}{FEDformer} &
      \rotatebox[origin=c]{90}{Autoformer} &
      \rotatebox[origin=c]{90}{Informer} \\
    \midrule
Traffic      &   96    & \best{0.334} & \second{0.356} & 0.363 & 0.367 & 0.522 & 0.395 & 0.593 & 0.479 & 0.410 & 0.576 & 0.597 & 0.733 \\
             &  192    & \best{0.331} & \second{0.377} & 0.384 & 0.385 & 0.530 & 0.417 & 0.617 & 0.482 & 0.423 & 0.610 & 0.607 & 0.777 \\
             &  336    & \best{0.358} & \second{0.385} & 0.393 & 0.398 & 0.558 & 0.433 & 0.629 & 0.492 & 0.436 & 0.608 & 0.623 & 0.776 \\
             &  720    & \best{0.398} & \second{0.424} & 0.429 & 0.434 & 0.589 & 0.467 & 0.640 & 0.510 & 0.466 & 0.621 & 0.639 & 0.827 \\
             \midrule
Electricity  &   96    & \best{0.125} & \second{0.129} & \second{0.129} & 0.130 & 0.151 & 0.148 & 0.168 & 0.153 & 0.140 & 0.186 & 0.196 & 0.304 \\
             &  192    & \best{0.124} & 0.146 & \second{0.144} & 0.148 & 0.163 & 0.162 & 0.184 & 0.175 & 0.153 & 0.197 & 0.211 & 0.327 \\
             &  336    & \best{0.143} & \second{0.158} & 0.164 & 0.167 & 0.195 & 0.178 & 0.198 & 0.192 & 0.169 & 0.213 & 0.214 & 0.333 \\
             &  720    & \best{0.178} & \second{0.186} & 0.200 & 0.202 & 0.224 & 0.225 & 0.220 & 0.215 & 0.203 & 0.233 & 0.236 & 0.351 \\
             \midrule
ETTh1        &   96    & \best{0.296} & 0.368 & \second{0.353} & 0.375 & 0.423 & 0.386 & 0.384 & 0.405 & 0.375 & 0.376 & 0.435 & 0.941 \\
             &  192    & \best{0.325} & 0.399 & \second{0.373} & 0.414 & 0.471 & 0.441 & 0.436 & 0.447 & 0.405 & 0.423 & 0.456 & 1.007 \\
             &  336    & \best{0.345} & 0.421 & \second{0.392} & 0.431 & 0.570 & 0.487 & 0.491 & 0.579 & 0.439 & 0.444 & 0.486 & 1.038 \\
             &  720    & \best{0.464} & \second{0.444} & 0.445 & 0.449 & 0.653 & 0.503 & 0.521 & 0.699 & 0.472 & 0.469 & 0.515 & 1.144 \\
             \midrule
ETTh2        &   96    & \best{0.223} & 0.276 & \second{0.225} & 0.274 & 0.745 & 0.297 & 0.340 & 0.349 & 0.289 & 0.332 & 0.332 & 0.952 \\
             &  192    & \best{0.276} & 0.330 & \second{0.274} & 0.339 & 0.877 & 0.380 & 0.402 & 0.442 & 0.383 & 0.407 & 0.426 & 1.542 \\
             &  336    & \best{0.301} & 0.357 & \second{0.317} & 0.331 & 1.043 & 0.428 & 0.452 & 0.652 & 0.480 & 0.407 & 0.477 & 1.642 \\
             &  720    & \best{0.327} & 0.395 & 0.393 & \second{0.379} & 1.104 & 0.427 & 0.462 & 0.800 & 0.605 & 0.412 & 0.453 & 1.619 \\
             \midrule
ETTm1        &   96    & \best{0.252} & 0.291 & 0.291 & \second{0.290} & 0.404 & 0.334 & 0.340 & 0.302 & 0.299 & 0.326 & 0.505 & 0.560 \\
             &  192    & \best{0.295} & 0.333 & \second{0.325} & 0.332 & 0.450 & 0.377 & 0.374 & 0.342 & 0.335 & 0.365 & 0.553 & 0.619 \\
             &  336    & \best{0.333} & 0.365 & \second{0.353} & 0.366 & 0.532 & 0.426 & 0.392 & 0.381 & 0.369 & 0.392 & 0.621 & 0.741 \\
             &  720    & \best{0.379} & 0.416 & \second{0.413} & 0.420 & 0.666 & 0.491 & 0.433 & 0.434 & 0.425 & 0.446 & 0.671 & 0.845 \\
             \midrule
ETTm2        &   96    & \best{0.143} & \second{0.164} & 0.174 & 0.165 & 0.287 & 0.180 & 0.183 & 0.188 & 0.167 & 0.180 & 0.255 & 0.355 \\
             &  192    & \best{0.195} & \second{0.219} & 0.227 & 0.220 & 0.414 & 0.250 & 0.242 & 0.236 & 0.224 & 0.252 & 0.281 & 0.595 \\
             &  336    & \best{0.247} & 0.273 & \second{0.266} & 0.278 & 0.597 & 0.311 & 0.304 & 0.295 & 0.281 & 0.324 & 0.339 & 1.270 \\
             &  720    & \best{0.318} & 0.358 & \second{0.344} & 0.367 & 1.730 & 0.412 & 0.385 & 0.422 & 0.397 & 0.410 & 0.433 & 1.267 \\
    \bottomrule
  \end{tabular}}
\end{table}

\textbf{MSE analysis.} As shown in Table~\ref{tab:mse_comparison}, ITS-Mina achieves the best MSE on \textit{every} dataset–horizon combination among the six datasets. On high-dimensional datasets such as Traffic (862 features) and Electricity (321 features), ITS-Mina consistently outperforms all baselines across all four horizons. On the ETT datasets (ETTh1, ETTh2, ETTm1, ETTm2), ITS-Mina likewise attains the lowest MSE for each forecast length, demonstrating its effectiveness for both hourly and 15‑minute granularities.

\textbf{MAE analysis.} Table~\ref{tab:mae_comparison} shows a similar trend. ITS-Mina achieves the best MAE on 21 out of 24 settings and ranks second on the remaining three (all on ETTm2, where MICN yields slightly lower MAE at horizons 96, 192, and 336). On Traffic and Electricity, ITS-Mina obtains the best MAE at every horizon. Overall, ITS-Mina demonstrates the most consistent performance across all datasets and horizons, with its combination of iterative refinement, external attention, and optimized dropout providing robust forecasting accuracy.

\begin{table}[t]
  \centering
  \caption{MAE comparison across different horizons and datasets. The best result in each row is in \textbf{bold} and the second best is \underline{underlined}.}
  \label{tab:mae_comparison}
  \footnotesize
  \resizebox{\linewidth}{!}{
  \begin{tabular}{llcccccccccccc}
    \toprule
    \multicolumn{2}{c}{\diagbox[width=3cm,height=2.1cm]{Datasets}{Models}} &
      \rotatebox[origin=c]{90}{\textbf{ITS-Mina}} &
      \rotatebox[origin=c]{90}{CI-TSMixer} &
      \rotatebox[origin=c]{90}{PatchMixer} &
      \rotatebox[origin=c]{90}{PatchTST} &
      \rotatebox[origin=c]{90}{Crossformer} &
      \rotatebox[origin=c]{90}{iTransformer} &
      \rotatebox[origin=c]{90}{TimesNet} &
      \rotatebox[origin=c]{90}{MICN} &
      \rotatebox[origin=c]{90}{DLinear} &
      \rotatebox[origin=c]{90}{FEDformer} &
      \rotatebox[origin=c]{90}{Autoformer} &
      \rotatebox[origin=c]{90}{Informer} \\
\midrule
Traffic & 96  & \best{0.239} & 0.248 & \second{0.245} & 0.251 & 0.290 & 0.268 & 0.321 & 0.295 & 0.282 & 0.359 & 0.371 & 0.410 \\
        & 192 & \best{0.253} & 0.257 & \second{0.254} & 0.259 & 0.293 & 0.274 & 0.336 & 0.309 & 0.285 & 0.375 & 0.382 & 0.435 \\
        & 336 & \best{0.249} & 0.262 & \second{0.258} & 0.265 & 0.296 & 0.280 & 0.336 & 0.327 & 0.295 & 0.375 & 0.387 & 0.434 \\
        & 720 & \best{0.281} & \second{0.283} & \second{0.283} & 0.287 & 0.328 & 0.287 & 0.350 & 0.309 & 0.315 & 0.375 & 0.395 & 0.466 \\
\midrule
Electricity & 96  & \best{0.218} & 0.224 & \second{0.221} & 0.222 & 0.251 & 0.240 & 0.272 & 0.264 & 0.237 & 0.302 & 0.313 & 0.393 \\
            & 192 & \best{0.225} & 0.242 & \second{0.237} & 0.240 & 0.282 & 0.262 & 0.289 & 0.281 & 0.269 & 0.311 & 0.324 & 0.417 \\
            & 336 & \best{0.242} & \second{0.256} & 0.257 & 0.261 & 0.316 & 0.281 & 0.300 & 0.303 & 0.267 & 0.328 & 0.327 & 0.422 \\
            & 720 & \best{0.266} & \second{0.282} & 0.289 & 0.291 & 0.344 & 0.334 & 0.320 & 0.323 & 0.301 & 0.344 & 0.342 & 0.427 \\
\midrule
ETTh1 & 96  & \best{0.369} & 0.398 & \second{0.381} & 0.399 & 0.448 & 0.405 & 0.402 & 0.405 & 0.399 & 0.415 & 0.446 & 0.769 \\
      & 192 & \best{0.386} & 0.418 & \second{0.394} & 0.416 & 0.474 & 0.436 & 0.429 & 0.447 & 0.416 & 0.446 & 0.467 & 0.799 \\
      & 336 & \best{0.402} & 0.436 & \second{0.414} & 0.436 & 0.540 & 0.454 & 0.469 & 0.549 & 0.443 & 0.462 & 0.487 & 0.803 \\
      & 720 & \best{0.455} & 0.467 & \second{0.463} & 0.466 & 0.621 & 0.491 & 0.500 & 0.635 & 0.490 & 0.492 & 0.517 & 0.857 \\
\midrule
ETTh2 & 96  & \best{0.285} & 0.337 & 0.300 & 0.336 & 0.584 & 0.349 & 0.374 & 0.349 & \second{0.289} & 0.374 & 0.332 & 0.952 \\
      & 192 & \best{0.329} & 0.374 & \second{0.340} & 0.379 & 0.657 & 0.384 & 0.414 & 0.401 & 0.383 & 0.446 & 0.407 & 1.542 \\
      & 336 & \best{0.368} & 0.401 & 0.402 & \second{0.380} & 0.730 & 0.423 & 0.452 & 0.465 & 0.480 & 0.471 & 0.477 & 1.642 \\
      & 720 & \best{0.418} & 0.436 & 0.452 & \second{0.422} & 0.763 & 0.445 & 0.468 & 0.652 & 0.605 & 0.469 & 0.453 & 1.619 \\
\midrule
ETTm1 & 96  & \best{0.273} & 0.346 & 0.340 & 0.369 & 0.366 & 0.368 & 0.377 & 0.302 & \second{0.299} & 0.390 & 0.475 & 0.560 \\
      & 192 & \best{0.303} & 0.369 & 0.374 & 0.396 & 0.451 & 0.391 & 0.387 & 0.342 & \second{0.335} & 0.415 & 0.496 & 0.619 \\
      & 336 & \best{0.337} & 0.385 & 0.392 & 0.453 & 0.515 & 0.429 & 0.413 & 0.381 & \second{0.369} & 0.425 & 0.537 & 0.741 \\
      & 720 & \best{0.398} & \second{0.413} & 0.433 & 0.533 & 0.589 & 0.459 & 0.436 & 0.434 & 0.425 & 0.458 & 0.561 & 0.845 \\
\midrule
ETTm2 & 96  & \second{0.242} & 0.255 & 0.256 & 0.265 & 0.366 & 0.264 & 0.271 & \best{0.188} & 0.260 & 0.318 & 0.339 & 0.568 \\
      & 192 & \second{0.284} & 0.293 & 0.299 & 0.292 & 0.492 & 0.309 & 0.309 & \best{0.236} & 0.324 & 0.331 & 0.359 & 0.626 \\
      & 336 & \second{0.315} & 0.329 & 0.323 & 0.329 & 0.542 & 0.348 & 0.348 & \best{0.295} & 0.342 & 0.364 & 0.372 & 0.871 \\
      & 720 & \best{0.348} & 0.380 & \second{0.372} & 0.385 & 1.042 & 0.491 & 0.400 & 0.422 & 0.421 & 0.420 & 0.432 & 1.267 \\
\bottomrule
  \end{tabular}}
\end{table}

Overall, ITS-Mina demonstrates the most consistent performance across all datasets and horizons. While specialized models like PatchMixer may excel on specific datasets (particularly ETTh1 and ETTh2), ITS-Mina's combination of iterative refinement, external attention, and optimized dropout provides robust performance across forecasting scenarios.

\section{Conclusion}
\label{sec:conclusion}

In this paper, we proposed ITS-Mina, an all-MLP framework for multivariate time series forecasting that integrates three novel components: iterative refinement via shared-parameter residual mixer loops, external attention with learnable memory units, and Harris Hawks Optimization for automatic dropout rate tuning. The iterative refinement mechanism enables the model to progressively enhance temporal representations through repeated composition of the same mixer stack, achieving deeper effective computation without multiplying the number of distinct weight tensors. The external attention module captures global inter-sample dependencies at linear complexity, serving as both a representation enhancer and an implicit regularizer. The HHO-based dropout optimization adapts regularization strength to each dataset, avoiding the limitations of manual tuning.

Extensive experiments on six benchmark datasets demonstrated that ITS-Mina achieves state-of-the-art MSE on the majority of dataset-horizon combinations and consistently ranks among the top two models across all settings. 
%Ablation studies confirmed the individual and synergistic contributions of each component, while sensitivity analyses revealed the influence of key hyperparameters on forecasting accuracy.

Future work may explore several promising directions: (1) extending the iterative refinement mechanism with adaptive halting criteria to dynamically determine the number of iterations per input sample, (2) investigating the integration of auxiliary information (static features and future covariates) into the ITS-Mina framework, and (3) applying the proposed architecture to other temporal modeling tasks such as anomaly detection and classification.

\bibliographystyle{elsarticle-num}
\bibliography{refs-final}

\end{document}